\documentclass[10pt,twocolumn,letterpaper]{article}

\usepackage{cvpr}
\usepackage{times}
\usepackage{epsfig}
\usepackage{graphicx}
\usepackage{amsmath}
\usepackage{amssymb}

\usepackage{subfigure}
\usepackage{algorithm}
\usepackage{algorithmic}
\usepackage{multirow}
\usepackage{wasysym}

\usepackage[breaklinks=true,bookmarks=false]{hyperref}

\cvprfinalcopy 

\def\cvprPaperID{808} 

\def\eg{\emph{e.g.}} 
\def\ie{\emph{i.e.}} 
\def\etal{\emph{et al.}}

\ifcvprfinal\fi
\begin{document}

\title{MRF Optimization by Graph Approximation}

\author{Wonsik Kim\thanks{Currently at Samsung Electronics}~~and Kyoung Mu Lee\\
Department of ECE, ASRI, Seoul National University, 151-742, Seoul, Korea\\
{\tt\small \{ultra16, kyoungmu\}@snu.ac.kr, \url{http://cv.snu.ac.kr}}
}

\maketitle
\thispagestyle{empty}

\begin{abstract}
Graph cuts-based algorithms have achieved great success in energy minimization for many computer vision applications. These algorithms provide approximated
solutions for multi-label energy functions via move-making approach. This approach fuses the current solution with a proposal to generate a lower-energy solution. Thus, generating the appropriate proposals is necessary for the success of the move-making approach. However, not much research efforts has been done on the generation of ``good'' proposals, especially for non-metric energy functions. In this paper, we propose an application-independent and energy-based approach to generate ``good'' proposals. With these proposals, we present a graph cuts-based move-making algorithm called GA-fusion (fusion with graph approximation-based proposals). Extensive experiments support that our proposal generation is effective across different classes of energy functions. The proposed algorithm outperforms others both on real and synthetic problems.
\end{abstract}

\vspace{-3mm}
\section{Introduction}
\label{sec:intro}
\vspace{-1mm}
Markov random field (MRF) has been used for numerous areas in computer
vision~\cite{PAMI/Szeliski08}.
MRFs are generally formulated as follows. Given a graph
$G=(\mathcal{V},\mathcal{E})$, the energy function of the pairwise MRF is given by

\begin{equation}
    E(\mathbf{x}) = \sum_{p \in \mathcal{V}} \theta_p (x_p) +
    \lambda \sum_{(p, q) \in \mathcal{E}} \theta_{pq} (x_p,x_q),
\label{eq:energy}
\end{equation}
where $\mathcal{V}$ is the set of nodes, $\mathcal{E}$ is the set
of edges, $x_p\in\{1,2,\cdots,L\}$ is the label assigned on node $p$, and $\lambda$ is the weight factor between unary and pairwise terms. Optimization of the MRF model is challenging because finding the global minimum of the energy function~(\ref{eq:energy}) is NP-hard in general cases.

There have been numerous researches on optimizing aforementioned function. Although they have been successful for many different applications, they still end up with unsatisfactory solutions when it comes to extremely difficult problems. In those kind of problems, many graph cuts-based algorithms cannot label sufficient number of nodes due to the strong non-submodularity and dual decomposition cannot decrease gaps due to many frustrated cycles~\cite{sontag2012efficiently}. In this paper, we address this problem by introducing simple graph cuts-based algorithm with the right choice of proposal generation scheme.

Graph cuts-based algorithms have attracted much attention as an optimization method for MRFs~\cite{PAMI/Kolmogorov04,PAMI/Boykov04,PAMI/Boykov01,Alahari2008/CVPR,Kohli2005/ICCV}.
Graph cuts can obtain the exact solution in polynomial time when the energy function~(\ref{eq:energy}) is submodular \cite{DAM/Boros02}. Even if the function is not submodular, a partial solution can be obtained with unlabeled nodes using roof duality (QPBO)~\cite{MP/Hammer84,CVPR/Rother07}. Graph cuts have also been used to solve multi-label energy functions. For this purpose, move-making algorithms have been proposed \cite{PAMI/Boykov01}, in which graph cuts optimize a sequence of binary functions to make moves.

\begin{figure}[t]
\begin{center}
    \mbox{%
    \includegraphics[width=0.9\linewidth]{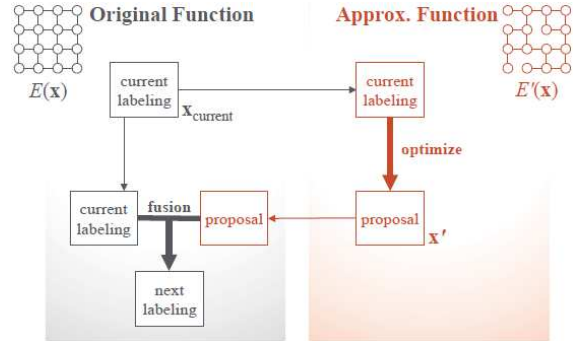}
    }
\end{center}
\vspace{-3mm}
\caption{The basic idea of the overall algorithm. The original function is approximated via graph approximation. The approximated function is optimized, and the solution is used as a proposal for the original problem.
}
\label{fig:overall}
\end{figure}

In a move-making algorithm, the most important decision is the choice of appropriate move-spaces. For example, in $\alpha$-expansion\footnote{In this paper, $\alpha$-expansion always refers to QPBO-based $\alpha$-expansion unless noted otherwise.}, move-spaces are determined by the selected $\alpha$ value. Simple $\alpha$-expansion strategy has obtained satisfactory results when the energy function is metric. Recently, $\alpha$-expansion has been shown to improve when the proper order of move-space $\alpha$ is selected instead of iterating a pre-specified order~\cite{Batra/CVPR11}.

However, $\alpha$-expansion does not work well when the energy function is non-metric. In such a case, reduced binary problems are no longer submodular. Performance is severely degraded when QPBO leaves a considerable number of unlabeled nodes. To solve this challenge, we need more elaborate proposals rather than considering homogeneous proposals as in $\alpha$-expansion. Fusion move~\cite{Lempitsky/PAMI10} can be applied to consider general proposals.

For the success of fusion algorithm, generating appropriate proposals is necessary. Although there has been a demand for a generic method of proposal generation~\cite{Lempitsky/PAMI10}, little research has been done on the mechanism of ``good'' proposal generation (we will specify the notion of ``good'' proposals in the next section). Instead, most research on proposal generation is often limited to application-specific approaches~\cite{CVPR/Woodford08,CVPR/Ishikawa09}.

In this paper, we propose a generic and application-independent approach to generate ``good'' proposals for non-submodular energy functions. With these proposals, we present a graph cuts-based move-making algorithm called GA-fusion (fusion with graph approximation-based proposals). This method is simple but powerful. It is applicable to any type of energy functions. The basic idea of our algorithm is presented in Figure~\ref{fig:overall}. Sec.~\ref{sec:prop_gen} and \ref{sec:algorithm} describes the algorithm in detail.

We test our approach both in non-metric and metric energy functions, while our main concern is optimizing non-metric functions. Sec.~\ref{sec:exp} demonstrates that the proposed approach significantly outperforms existing algorithms for non-metric functions and competitive with other state-of-the-art for metric functions. For the non-metric case, our algorithm is applied to image deconvolution and texture restoration in which conventional approaches often fail to
obtain viable solutions because of strong non-submodularity. We also evaluated our algorithm on synthetic problems to show robustness to the various types of energy functions.
\vspace{-1mm}
\section{Background and related works}
\label{sec:background}
\vspace{-1mm}
\subsection{Graph cuts-based move-making algorithm}
\vspace{-1mm}
Graph cuts-based algorithms have a long history~\cite{PAMI/Kolmogorov04,PAMI/Boykov04,PAMI/Boykov01,Alahari2008/CVPR,Kohli2005/ICCV}. These algorithms have extended the class of applicable energy functions from binary to multi-label, from metric to non-metric, and from pairwise to
higher-order energies (among these, higher-order energies are not the main concern of this paper).

Graph cuts can obtain the global minimum when the energy function~(\ref{eq:energy}) is submodular. In the binary case, a function is submodular if every pairwise term satisfies $\theta_{00}+\theta_{11} \leq \theta_{01}+\theta_{10}$, where $\theta_{00}$ represents $\theta_{pq}(0,0)$.


Graph cuts have also been successfully applied to multi-label
problems. One of the most popular schemes is $\alpha$-expansion.
$\alpha$-Expansion reduces optimization tasks into minimizing a
sequence of binary energy function

\begin{equation}
    E_b(\mathbf{y}) = E(\mathbf{x}_b(\mathbf{y})),
\label{eq:a-expansion}
\end{equation}
where $E_b(\mathbf{y})$ is the function of a binary vector
$\mathbf{y}\in\{0,1\}^{|\mathcal{V}|}$, and
$\mathbf{x}_b(\mathbf{y})$ is defined by

\begin{equation}
    x_{b,p}(y_p) = (1-y_p)\cdot x^{cur}_p + y_p \cdot \alpha,
\label{eq:auxiliary}
\end{equation}
where $x_{b,p}(y_p)$ is an element-wise operator for $\mathbf{x}_b(\mathbf{y})$ at node $p$, and $x^{cur}_p$ denotes the current label assigned on node $p$. The label on node $p$ switches between the current label and $\alpha$ according to the value of $y_p$. In such a case, the binary function $E_b(\mathbf{y})$ is submodular if the original function is metric~\cite{PAMI/Boykov01}. This condition is relaxed in~\cite{PAMI/Kolmogorov04} such that the binary function $E_b(\mathbf{y})$ is submodular if every pairwise term satisfies 

\begin{equation}
    \theta_{\alpha,\alpha}+\theta_{\beta,\gamma} \leq \theta_{\alpha,\gamma}+\theta_{\beta,\alpha}.
\label{eq:metric}
\end{equation}

$\alpha$-Expansion is one of the most acclaimed methodologies; however, standard $\alpha$-expansion is not applicable if the energy function does not satisfy the condition~(\ref{eq:metric}). In such a case, a sequence of reduced binary functions is no longer submodular. We may truncate the pairwise
terms~\cite{PAMI/Szeliski08,SIGGRAPH/Agarwala02} to optimize these functions, thereby making every
pairwise term submodular. This strategy works only when the non-submodular part of the energy function is very small. If the non-submodular part is not negligible, performance is seriously degraded~\cite{CVPR/Rother07}.

For the second option, QPBO-based $\alpha$-expansion can be used. In this approach, QPBO is used to optimize sub-problems of $\alpha$-expansion ($\ie$, reduced binary functions). QPBO gives optimal solutions for submodular binary functions; it is also applicable to non-submodular functions. For non-submodular
functions, however, QPBO leaves a certain number of unlabeled nodes. Although QPBO-based $\alpha$-expansion is usually considered as a better choice than the truncation, it also performs very poorly when the reduced binary functions have a strong non-submodularity, which creates numerous unlabeled nodes.

For the third option, QPBO-based fusion move can be considered~\cite{Lempitsky/PAMI10}. Fusion move is a
generalization of $\alpha$-expansion. It produces binary functions in a way similar with $\alpha$-expansion (Equation~(\ref{eq:a-expansion})). The only difference is the operator $x_{b,p}(y_p)$, which is defined as follows:

\begin{equation}
    x_{b,p}(y_p) = (1-y_p)\cdot x^{cur}_p + y_p \cdot x^{pro}_p,
\label{eq:auxiliary}
\end{equation}
where $x^{pro}_p$ is a proposal labeling at node $p$. The value of $x^{pro}_p$ can be different for each node contrary to the case in $\alpha$-expansion. In this case, the function $E_b(\mathbf{y})$ is not always guaranteed to be submodular.

\vspace{-1mm}
\subsection{Proposals for fusion approach}
\label{sec:proposal}
\vspace{-1mm}

\renewcommand{\multirowsetup}{\centering}
\begin{table}[t]
\caption{Four types of proposal generation strategies.}
\begin{center}
\footnotesize
\begin{tabular}{lcccc}
\hline
 & Online- & Energy- & \multirow{2}{*}{Generality} & \\
 & generation & awareness & & \\
\hline
\textit{Type 1} & -            & -            & -            & \cite{Lempitsky/PAMI10}\cite{CVPR/Woodford08} \\
\textit{Type 2} & $\checkmark$ & -            & -            & \cite{CVPR/Ishikawa09} \\
\textit{Type 3} & $\checkmark$ & $\checkmark$ & $\vartriangle$  & \cite{ICCV/Ishikawa09} \\
\textit{Type 4} & $\checkmark$ & $\checkmark$ & $\checkmark$ & Proposed \\

\hline
\end{tabular}
\end{center}
\label{table:types}
\end{table}

When the fusion approach is considered, the immediate concern is related to the generation of the proposals. The choice of proposals changes move-spaces as well as the difficulties of the sub-problems, by
changing the number of non-submodular terms, which consequently affects the qualities of the final solutions.

Although choosing appropriate proposals is of crucial importance, little research has been conducted on generating good proposals. Previous approaches can be roughly divided into two categories: offline and online generation. Most existing approaches generate proposals offline (\textit{type 1} in Table~\ref{table:types}). Before the optimization begins, multiple number of hypotheses are generated by some heuristics. For example, Woodford $\etal$~\cite{CVPR/Woodford08} used approximated disparity maps as proposals for stereo application. Lempitsky $\etal$~\cite{Lempitsky/PAMI10} used the Lucas-Kanade (LK) and the Horn-Schunck (HS) methods with various parameter settings for optical flow. They do not take the objective energy function into account when generating proposals. Also, the number of proposals is limited by predetermined parameters. In addition, those proposals are application-specific and require domain knowledge.

Some other approaches generate proposals in runtime (\textit{type 2}, \textit{type 3}). Contrary to \textit{type 1}, the number of the proposals is not limited since they dynamically generate proposals online. In~\cite{CVPR/Ishikawa09}, proposals are generated by blurring the current solution and random labeling for denoising application. However, they do not explicitly concern objective energy in proposal generation. And, they are also application-specific. Recently, Ishikawa~\cite{ICCV/Ishikawa09} proposed an application-independent method to generate proposals. This method uses gradient descent algorithm on the objective energy function. Although they are energy-aware and can be applied to some cases, it is still limited to differentiable energy functions. Thus, this method cannot be applied even to the Potts model, which is one of the most popular prior models. In our understanding, this algorithm is only meaningful for ordered labels that represent physical quantities.

We introduce a new type of proposal generation (\textit{type 4}). Proposals dynamically generated online so that the number of the proposals is not limited, unlike \textit{type 1}, which uses pre-generated set of proposals. The proposals are generated in the energy-aware way so that the energies of proposals have obvious correlation with final solution. In addition, it is generic and applicable to any class of energy functions.

Lempitsky $\etal$ pointed out two properties for ``good'' proposals: \textit{quality} of individual proposal and \textit{diversity} among different proposals~\cite{Lempitsky/PAMI10}. In addition, we claim in this paper that \textit{labeling rate} is another important factor in measuring the quality of a proposal.

The three properties for good proposals are summarized in follows:

\begin{itemize}
    \item   {\bf Quality} Good proposals are close to minimum such that proposals can guide the solution to minimum by fusion moves. In other words, good proposals have low energy.
    \item   {\bf Diversity} For the success of the fusion approach, diversity among different proposals is required.
    \item   {\bf Labeling rate} Good proposals result in high labeling rate when they are fused with the current solution. In other words, good proposals produce easy-to-solve sub-problems.
\end{itemize}

Note that these conditions are not always necessary. One may think of proposals that do not meet the foregoing conditions, but help to obtain a good solution. However, in general, if proposals satisfy
these conditions, we can expect to obtain a good solution. In Sec.~\ref{sec:exp}, we empirically show that our proposal exhibits the above properties.

\vspace{-1mm}
\section{Proposal generation via graph approximation}
\label{sec:prop_gen}
\vspace{-1mm}

\subsection{Graph approximation}
\label{sec:edge}
\vspace{-1mm}

We approximate the original objective function (\ref{eq:energy}) to relieve difficulties from non-submodularity. Our motivation comes from the well-known fact that less connectivity
of a graph makes fewer unlabeled nodes~\cite{CVPR/Rother07}.

We exploit graph approximation by edge deletion to obtain an approximated function. This approximation is applicable to any class of energy functions, yet they are simple and easy. In graph approximation, a graph $G=(\mathcal{V},\mathcal{E})$ is approximated as $G'=(\mathcal{V},\mathcal{E}')$.

More specifically, we approximate the original graph with a random subset $\mathcal{E}'$ of edges from the original edge set $\mathcal{E}$. Pairwise terms $\theta_{pq}$, where $(p,q) \in \mathcal{E} \backslash \mathcal{E}'$, are dropped from the energy formulation~(\ref{eq:energy}). The approximated function is given by the following.

\begin{equation}
    E'(\mathbf{x}) = \sum_{p \in \mathcal{V}} \theta_p (x_p) +
    \lambda \sum_{(p, q) \in \mathcal{E}'} \theta_{pq} (x_p,x_q).
\label{eq:approx}
\end{equation}

To achieve three properties for ``good proposals'' mentioned in Sec. \ref{sec:proposal}, two conditions are required for an approximated function $E'(\mathbf{x})$. First, the approximated function should be easy to solve although the original one $E(\mathbf{x})$ is difficult. In other words, more nodes are labeled when we apply simple $\alpha$-expansion algorithm. Second, the approximated function should be similar to the original one. In other words, solution $\mathbf{x}'$ of
the approximated function should have low energy in terms of the original function. Those characteristics are examined in next section.

There have been other approaches to approximate the original function in restricted structures. Some structure are known to be tractable, such as bounded treewidth subgraphs ($\eg$ tree and outer-planar graph)~\cite{IT/Wainwright05,PAMI/Kolmogorov06,CVPR/veksler05,CVPR/batra10}. However, our approximation is not restricted to any type of special structure.

The inappropriateness of these structured approximations to our framework can be attributed to two main reasons. First, the approximation with the restricted structures requires the deletion of too many edges. For example, tree structures only have $|\mathcal{V}|-1$ edges, and 2-bounded treewidth graphs have at most $2|\mathcal{V}|-3$ edges. In practice, the number of edges are usually smaller than $2|\mathcal{V}|-3$. It is not a desirable scenario particularly for highly connected graphs. Second, exact optimization of 2-bounded treewidth graphs requires too much time. Several seconds to tens of seconds may be needed on the moderate size of graphs typically used in computer vision~\cite{Fix/CVPR12,CVPR/batra10}. Therefore, embedding this structure to our iterative framework is not appropriate.

Recently, \cite{gorelick2014submodularization} proposesd the method which iteratively minimizes the approximated function. There are two main difference with ours. First, they approximate the energies in the principled way so that the approximation is same with the original one within the trust region or is an upper bound of the original one while ours merely drop the randomly chosen edges. Second, by careful approximation, they guarantee the solution always decreases the original energy while ours allow energy to increase in the intermediate step.

In the experimental section, we investigate the approximation with spanning trees and show that it severely degrades the performance.

\vspace{-1mm}
\subsection{Characteristics of approximated function}
\label{sec:characteristics}
\vspace{-1mm}

\begin{figure}[t]
\begin{center}
    \mbox{%
    \subfigure[]{ \includegraphics[width=0.5\linewidth]{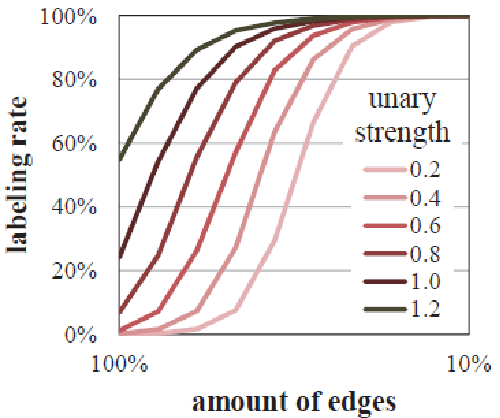}\label{fig:edgea}}
    \subfigure[]{ \includegraphics[width=0.5\linewidth]{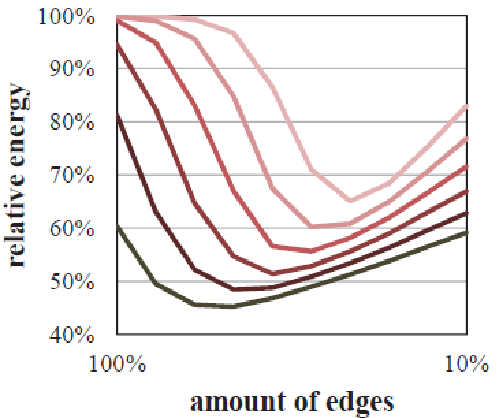}\label{fig:edgeb}}
    }
\end{center}
\vspace{-3mm}
\caption{(a) Labeling rates and (b) relative energies are depicted
as the graph is approximated with a random subset of edges. Relative energies are calculated with the original functions. With approximation, the labeling rate increases and the relative energy becomes lower.} \label{fig:edge}
\end{figure}

In this section, we experimentally show that the graph approximation
strategy achieves the two aforementioned conditions. Through the
approximation, solving the function becomes easier, and the solution of the approximation has low energy in terms of original
function.

We design the following experiments to meet the study objectives. First, we
build the binary non-submodular energy functions on a 30-by-30 grid graph with
4-neighborhood structure. Unary and pairwise costs are determined as follows.

\begin{equation}
    \theta_p (0) = 0, \theta_p (1) = k_p, \text{~~~or~~~} \theta_p (0) = k_p, \theta_p (1) = 0,
\label{eq:unary}
\end{equation}
\begin{equation}
    \theta_{pq} (x_p,x_q) =
    \begin{cases}
    0& \text{if $x_p=x_q$},\\
    s_{pq} \gamma_{pq}& \text{if $x_p \neq x_q$},
    \end{cases}
\label{eq:pair}
\end{equation}
where $k_p$ and $\gamma_{pq}$ are taken from a uniform
distribution $U(0,1)$, and $s_{pq}$ is randomly chosen from
$\{-1,+1\}$. When $s_{pq}$ is $+1$, the corresponding pairwise
term is metric. To vary the difficulties of the problems, we
control the unary strength, which is computed as
$\text{mean}_{p,i} \theta_p (i)/ \text{mean}_{p,q,i,j} \theta_{pq}
(i,j)$ after conversion into normal form. Since above energy function is already written in normal form, it is easy to set the desired unary strength by changing the
weight factor $\lambda$. The unary strength is changed from 0.2 to 1.2,
with interval of 0.2. For each unary strength, 100 random
instances of energy function were generated. As unary strength
decreases, QPBO produces more unlabeled nodes. Of all nodes, $54.7\%$ are labeled with the unary strength of 1.2, and none are labeled
with the unary strength of 0.2.

We approximate the foregoing functions by graph approximation and then optimize them using QPBO. For approximated functions, more nodes
are labeled than the original ones. The obtained solutions have
low energies in terms of original functions. These results\footnote{Here, relative
energy is given by the energy of the solution divided by the
energy of the labeling with zero for all nodes. The unlabeled
nodes in the solution are labeled with zero.} are
summarized in Figure~\ref{fig:edge}. When the approximation uses a
smaller subset $\mathcal{E}'$, more nodes are labeled. Those
results demonstrate that the proposed approximation makes the
problem not only easy to solve but also similar to the original
function.

\vspace{-1mm}
\section{Overall algorithm}
\label{sec:algorithm}
\vspace{-1mm}

The basic idea of the overall algorithm is depicted in
Figure~\ref{fig:overall}, which illustrates a single iteration of the
proposed algorithm. Our algorithm first approximates original target function and then optimizes it to generate proposals.

A single iteration of algorithm is composed of two steps: proposal generation and fusion, as presented in Algorithms \ref{alg:overall} and \ref{alg:exp_approx}. To generate proposals, we first obtain an approximated function $E'(\mathbf{x})$ of the original $E(\mathbf{x})$ with $\rho \cdot 100$ percent of edges.

Estimation of the optimal $\rho$ is not an easy task. As shown in Figure~\ref{fig:edge}, the minimum changes when the unary strength varies. We have tried two extremes to choose the parameter $\rho$. First, we simply fixed the $\rho$ value throughout all the iteration. It did not work since the optimal $\rho$ value changed not only for each problem, but also for each iteration. And then, we tried to estimate the optimal $\rho$ value every time. It also turned out to be inefficient because it caused too much overhead in time. Instead of taking one of these two extreme approaches, the parameter $\rho$ is randomly drawn from the uniform distribution $U(0,1)$ for each iteration. This simple idea gives surprisingly good performance in the experiments.

Having an approximated function $E'(\mathbf{x})$, we perform $K$ iterations of $\alpha$-expansion using the current labeling $\mathbf{x}_\text{current}$ as the initial solution. Solution $\mathbf{x}'$ obtained by optimizing the approximated function is then fused with $\mathbf{x}_\text{current}$. Note that, the approximated function $E'(\mathbf{x})$ is not fixed throughout the entire procedure, but it dynamically changes to give diversity to proposals. Larger $K$ tends to produce poor proposals by drift the solution far from the current state while too small $K$ tends to produce proposals similar to current state. The iteration $K$ is set to be $\min(5,L)$ throughout entire experiments. At line 6 in Alg. \ref{alg:exp_approx}, the minimum of $E_b'$ is not always tractable due to the non-submodularity. We approximate the minimum by QPBO while fixing unlabeled node to zero.

\begin{algorithm}[t]
\small
\caption{GA-fusion algorithm}
\label{alg:overall}
\begin{algorithmic}[1]

\STATE initialize the solution $\mathbf{x}_\text{current}$

\REPEAT

\STATE \textbf{$<$proposal generation$>$}
\STATE $\mathbf{x}_\text{proposal}\leftarrow \text{OptimizeGA} (\mathbf{x}_\text{current})$

\STATE \textbf{$<$fusion$>$}
\STATE $\mathbf{x}_\text{current} \leftarrow \text{FUSE}(\mathbf{x}_\text{current}, \mathbf{x}_\text{proposal})$

\UNTIL{the algorithm converges.}
\end{algorithmic}
\end{algorithm}

\begin{algorithm}[t]
\small
\caption{$\text{OptimizeGA}(\mathbf{x})$}
\label{alg:exp_approx}
\begin{algorithmic}[1]

\STATE initialize the solution with $\mathbf{x}$
\FOR{$i = 1 \to K$}

\STATE build a binary function $E_b$ for expansion with the label $\alpha$
\STATE $\rho \sim U(0,1)$
\STATE approximate $E_b$ by $E_b'$ using $\rho \times 100$ percent of randomly chosen edges
\STATE $\mathbf{x} \leftarrow \arg \min_\mathbf{x} E_b'$

\ENDFOR

\RETURN $\mathbf{x}$

\end{algorithmic}
\end{algorithm}

\vspace{-1 mm}
\section{Experiments}
\label{sec:exp}
\vspace{-1mm}

\vspace{-1mm}
\subsection{Non-metric function optimization}
\vspace{-1mm}
\subsubsection{Image deconvolution}
\label{sec:deconvolution}
\vspace{-1mm}

\begin{figure}[t]
\begin{center}
    \mbox{%
    \includegraphics[width=0.9\linewidth]{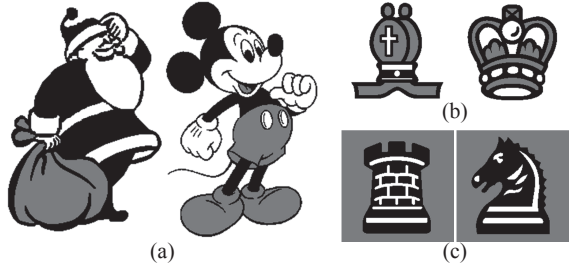}
    }
\end{center}
\vspace{-3mm}
\caption{Example input images of deconvolution from (a) 'characters', (b) 'white chessmen', and (c) 'black chessmen' datasets.
}
\label{fig:decon_input}
\end{figure}

Image deconvolution is the recovery of an image from a blurry and noisy image~\cite{Raj/CVPR05}. Given its high connectivity and strong non-submodularity, this problem has been reported as a challenging one~\cite{CVPR/Rother07}. The difficult nature of the problem particularly degrades the performance of graph cuts-based algorithms. In the benchmark~\cite{CVPR/Rother07}, graph cuts-based algorithms have achieved the poorest results. However, we demonstrate in the following that graph cuts-based algorithm can be severely improved by the proper choice of proposals.

For experiments, we construct the same MRF model used in~\cite{Raj/CVPR05}. First, the original image (colored with three labels) is blurred with $3 \times 3$ Gaussian kernel where $\sigma=3$. The image is again distorted with Gaussian pixel-wise noise with $\sigma=10$. For reconstruction, the MRF model with $5 \times 5$ neighborhood window is constructed. Smoothness is given by the Potts model.

We tested various algorithms on three datasets in Figure~\ref{fig:decon_input}. They include `characters' dataset (5 images, 200-by-320 pixels), `white chessmen' dataset (6 images, 200-by-200 pixels), and `black chessmen' dataset (6 images, 200-by-200 pixels)\footnote{Whole data set will be provided in the supplementary material}. We compare GA-fusion with other graph cuts-based algorithms. They only differ in the strategies to generate proposals: homogeneous labeling ($\alpha$-expansion), random labeling (random-fusion), dynamic programming on random spanning tree (ST-fusion), and proposed one (GA-fusion). The results imply that it is important to choose proper proposals. Note that truncation for non-submodular part \cite{rother2005digital} ($\alpha$-expansion(t)) degrades the performance.

We also apply other algorithms including belief propagation (BP)~\cite{ICCV/Tappen03,pearl1988probabilistic}, sequential tree-reweighted message passing (TRW-S)~\cite{IT/Wainwright05,PAMI/Kolmogorov06}, and max-product linear
programming with cycle and triplet (MPLP-C)~\cite{sontag2012efficiently}. For BP, TRW-S, and MPLP-C, we used source codes provided by the authors.

\begin{figure*}[t]
\begin{center}
    \mbox{%
    \subfigure[GA-fusion]{ \includegraphics[width=0.13\linewidth]{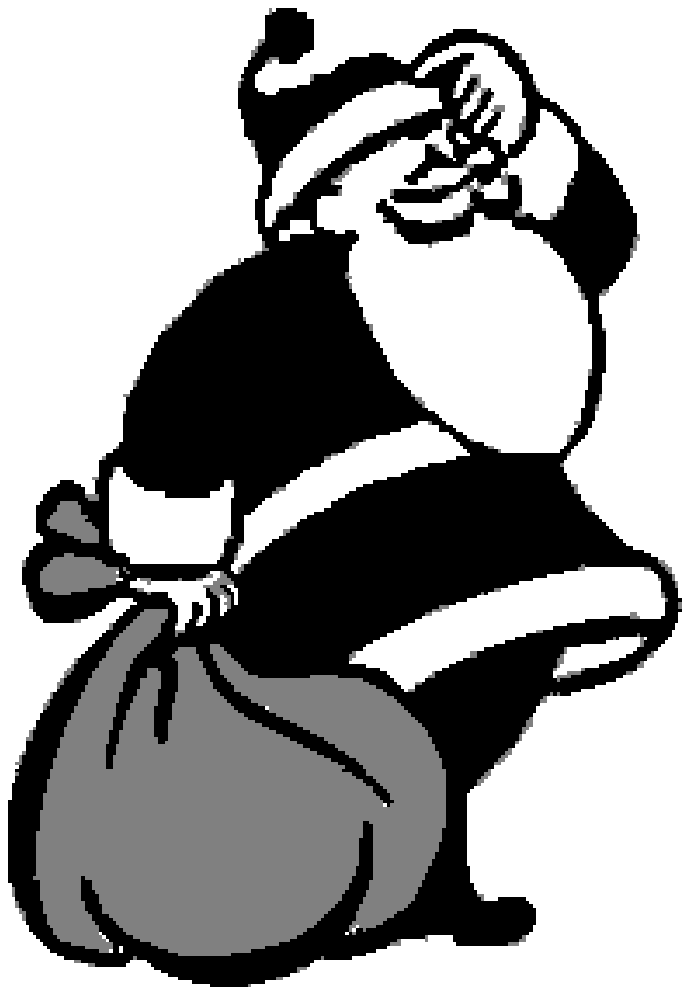}}
    \subfigure[ST-fusion]{ \includegraphics[width=0.13\linewidth]{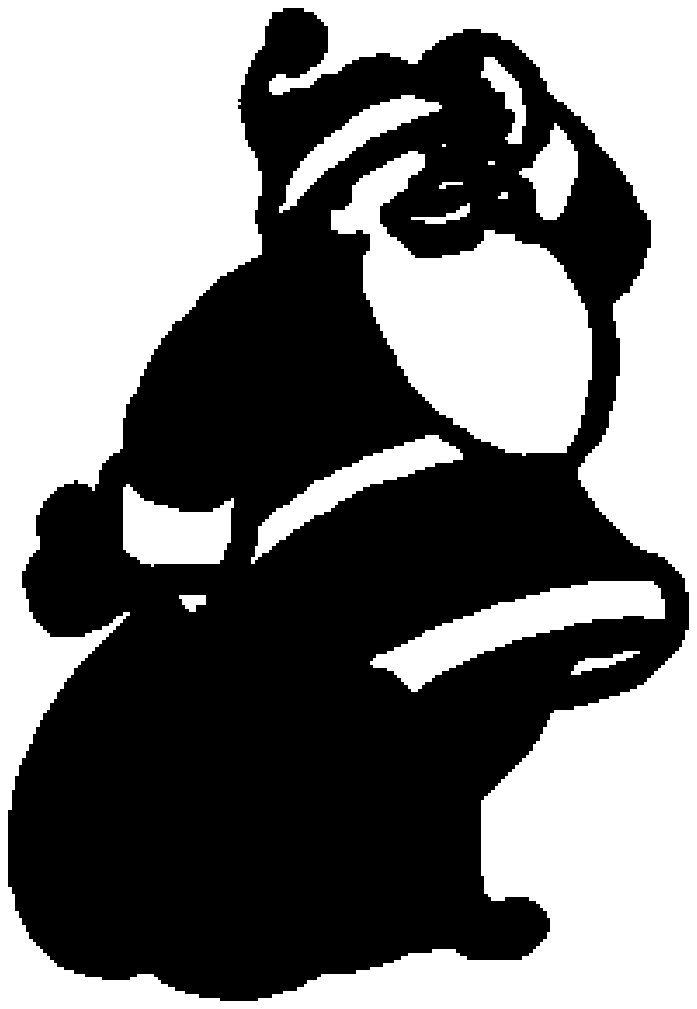}\label{fig:alpha}}
    \subfigure[$\alpha$-expansion]{ \includegraphics[width=0.13\linewidth]{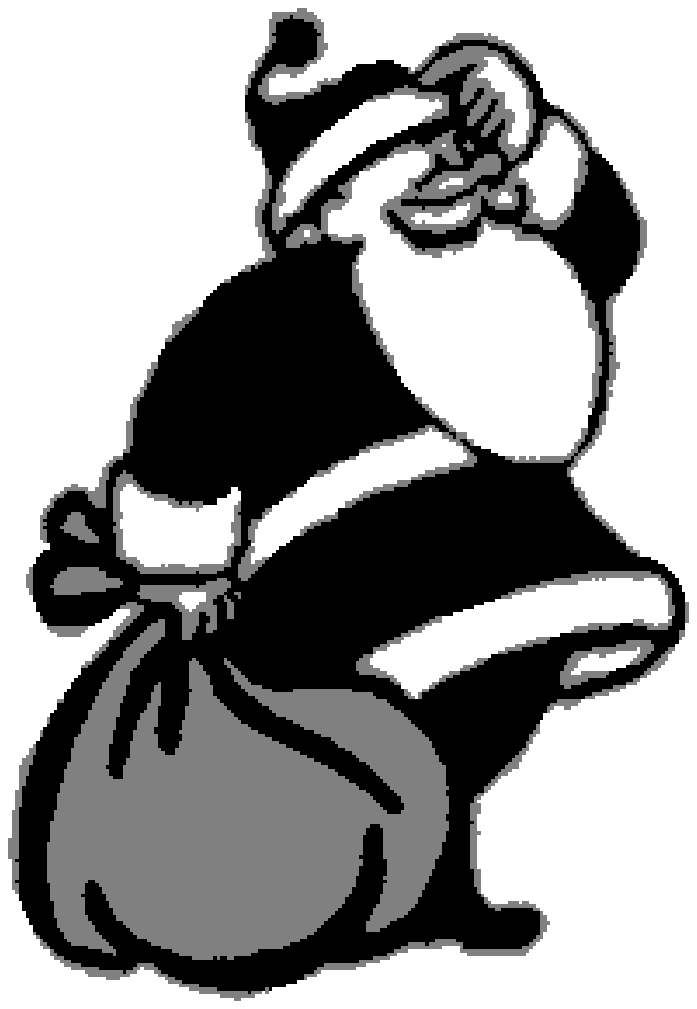}\label{fig:random}}
    \subfigure[Random-fusion]{ \includegraphics[width=0.13\linewidth]{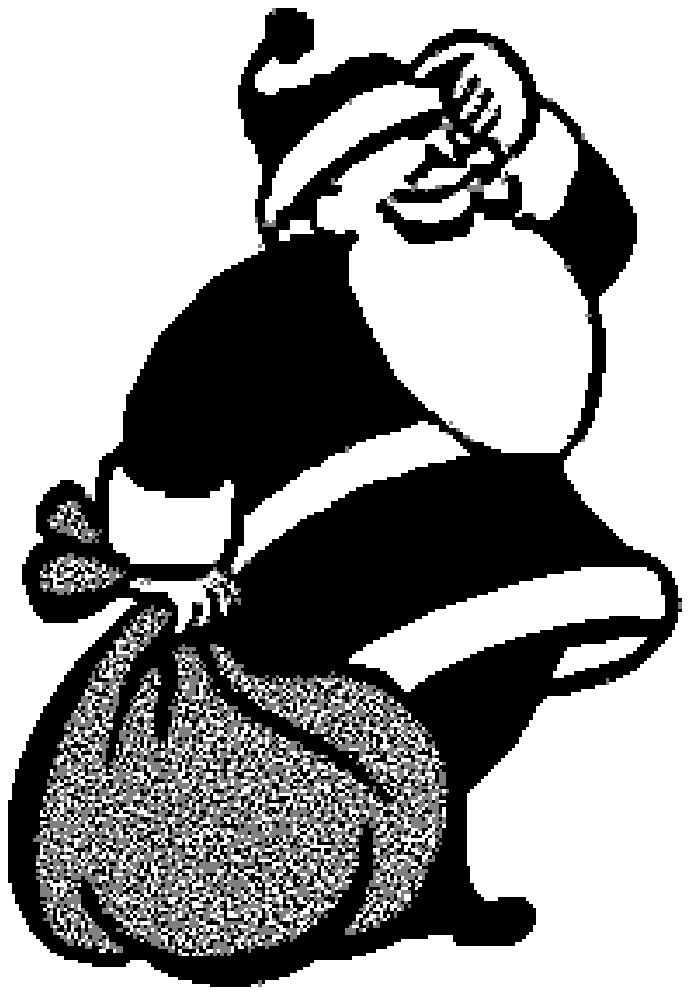}\label{fig:GA}}
    \subfigure[BP]{ \includegraphics[width=0.13\linewidth]{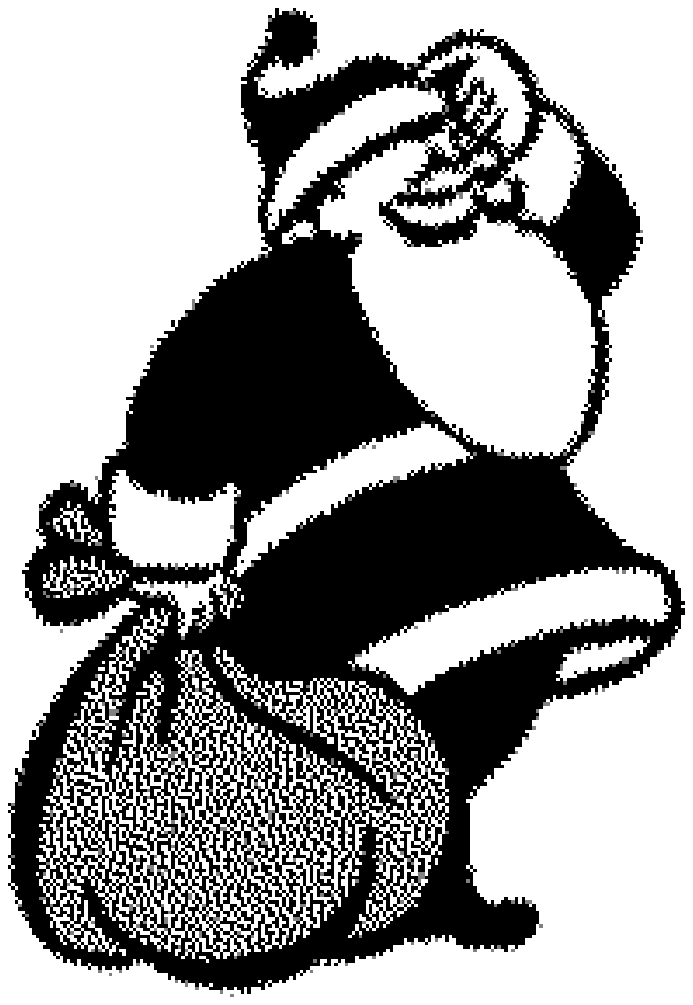}\label{fig:GA}}
    \subfigure[TRW]{ \includegraphics[width=0.13\linewidth]{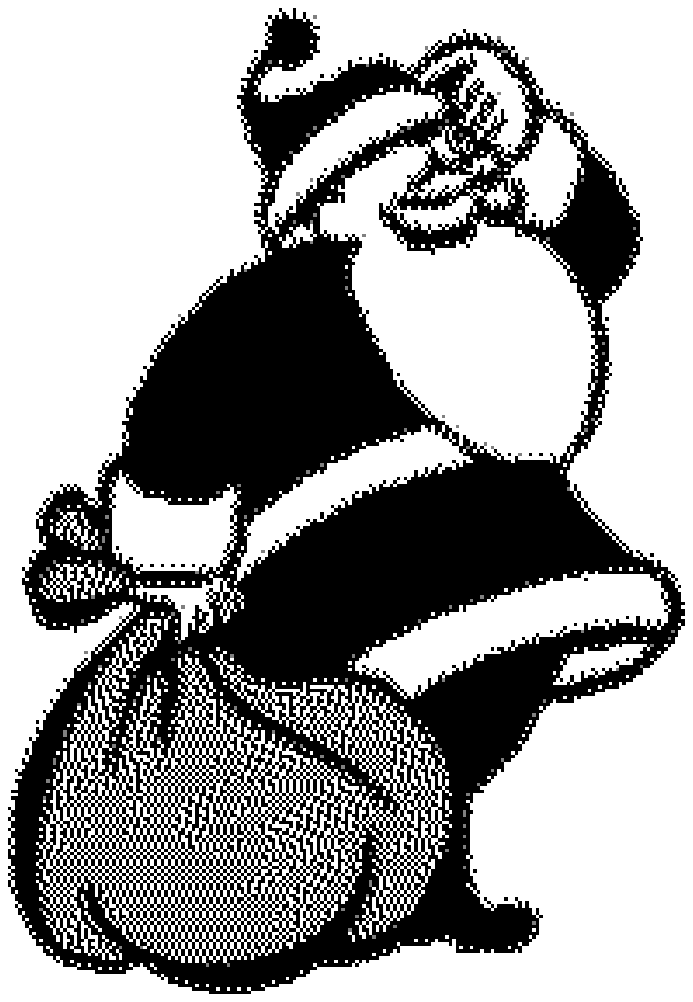}\label{fig:GA}}
    \subfigure[MPLP-C]{ \includegraphics[width=0.13\linewidth]{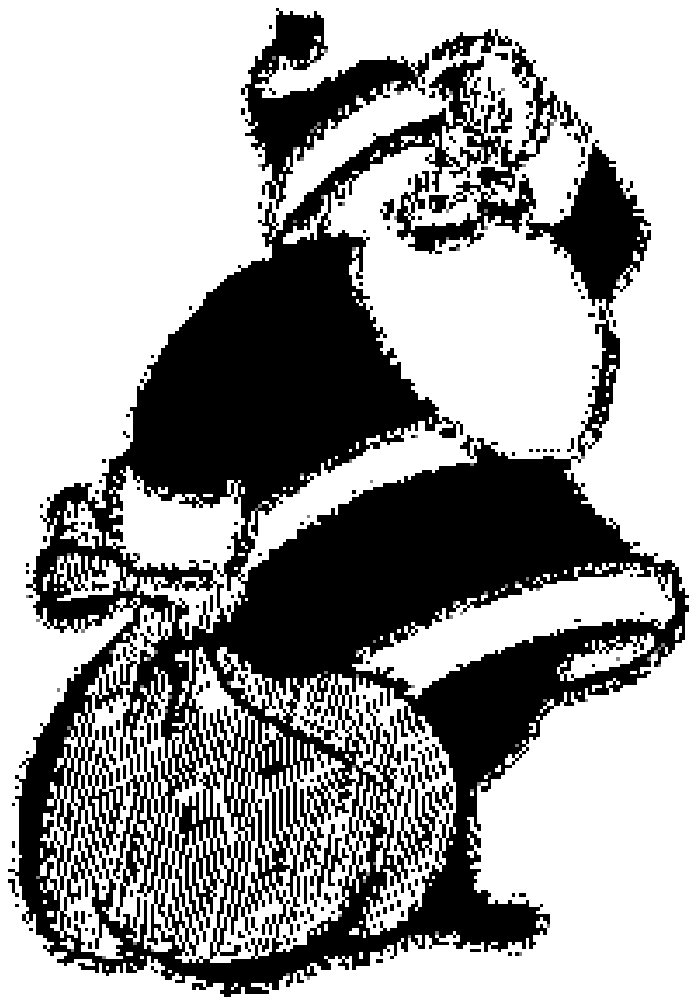}\label{fig:GA}}
    }
\end{center}
\vspace{-3mm}
\caption{Image deconvolution results on the Santa image. Proposed GA-fusion algorithm achieves best results. (a--d) Four graph cuts-based algorithms obtain significantly different results. It implies that the proper choice of proposal is crucial for the success of the graph cut-based algorithm.
}
\label{fig:decon_santa}
\end{figure*}

The results are summarized in Table~\ref{table:deconvolution}. GA-Fusion always achieves lowest energy solution. Figure~\ref{fig:decon_santa} shows quantitative results for the Santa image. Only GA-fusion achieved a fine result. $\alpha$-Expansion converged in 3.51 seconds on average. Other algorithms are iterated for 30 seconds except for MPLP-C, which is iterated for 600 seconds.

\begin{figure}[t]
\begin{center}
    \mbox{%
    \subfigure{ \includegraphics[width=0.37\linewidth]{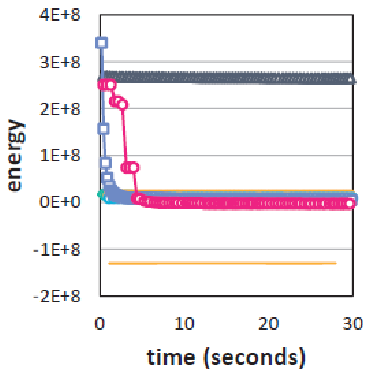}}
    \subfigure{ \includegraphics[width=0.37\linewidth]{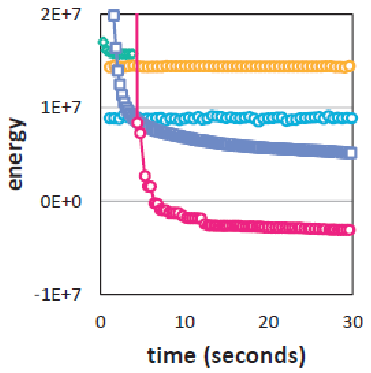}}
    \subfigure{ \includegraphics[width=0.19\linewidth]{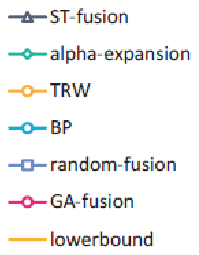}}
    }
\end{center}
\vspace{-3mm}
\caption{Energy decrease of each method for the deconvolution of the Santa image. Two plots shows the same curves from a single experiment, with different scales on the $y$-axis.
}
\label{fig:decon_timing}
\end{figure}

\begin{figure}[t]
\begin{center}
    \mbox{%
    \subfigure{ \includegraphics[width=0.5\linewidth]{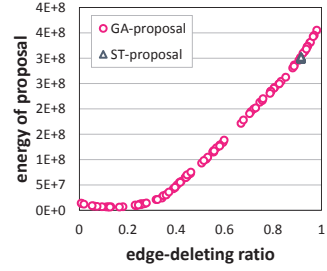}}
    }
\end{center}
\vspace{-3mm}
\caption{Original energy is approximated and optimized by two different methods (GA and ST). For each method 100 different random results are plotted. GA-proposals usually have lower energy than ST-proposals because random spanning tree approximation deletes too many edges.
}
\label{fig:decon_approx}
\end{figure}

\begin{figure}[t]
\begin{center}
    \mbox{%
    \subfigure{ \includegraphics[width=0.45\linewidth]{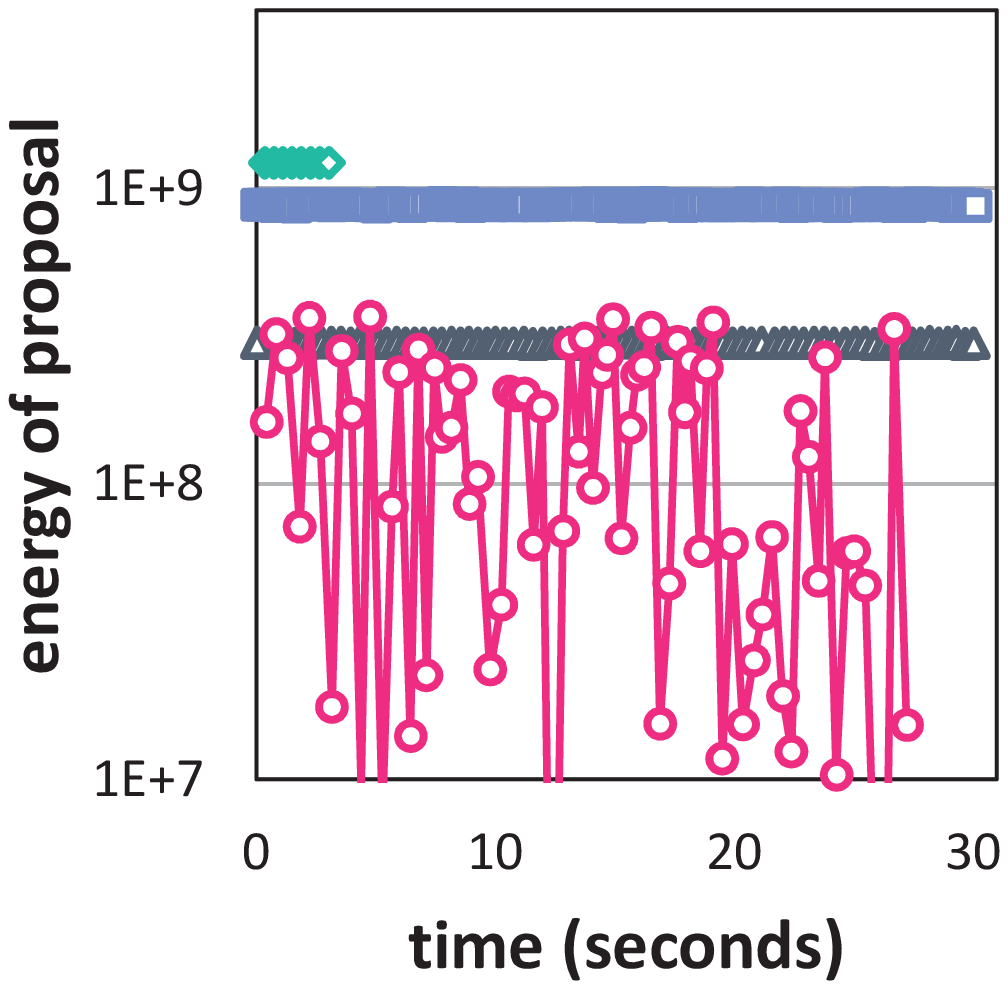}}
    \subfigure{ \includegraphics[width=0.45\linewidth]{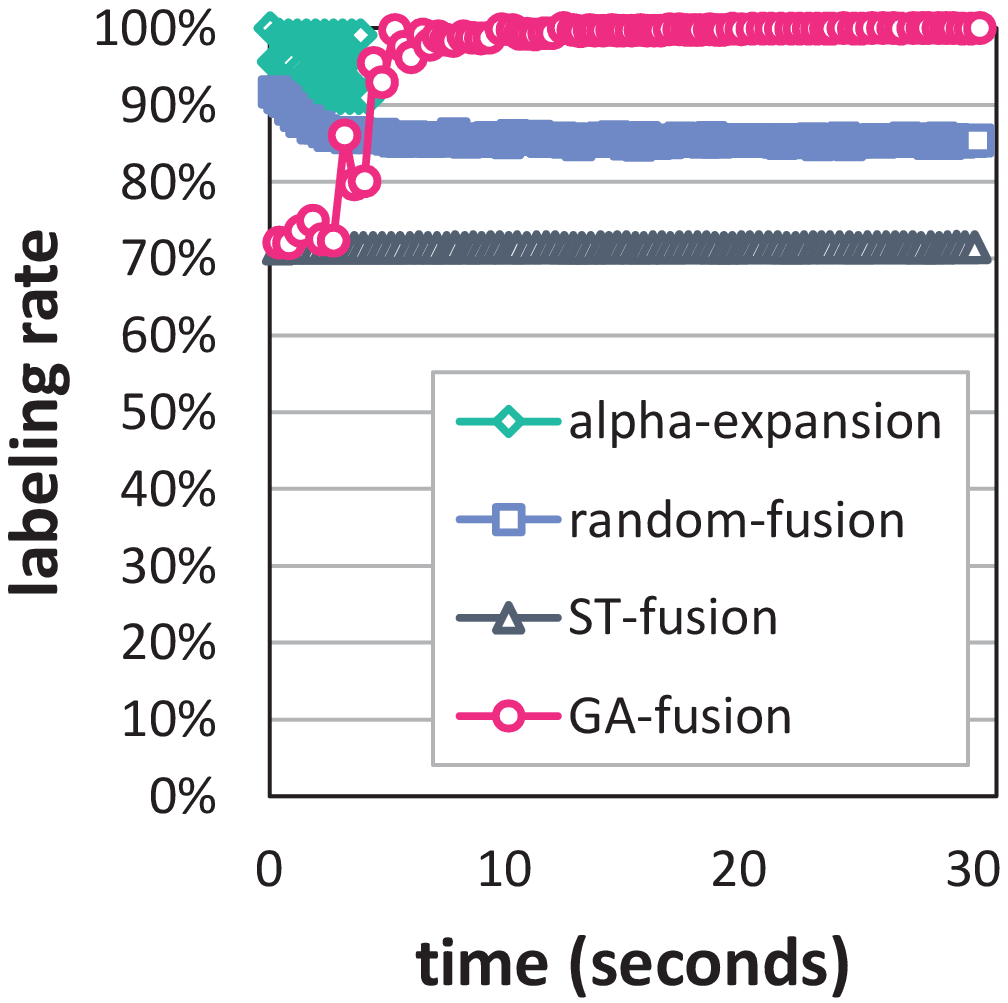}}
    }
\end{center}
\vspace{-3mm}
\caption{Experiment on deconvolution of the Santa image. (Left) Quality (energy) of the proposals for each iteration using a log scale. (Right) Labeling rate with the proposals for each iteration.
}
\label{fig:decon_proposal}
\end{figure}

We provide more detailed analysis with the Santa image in Figures~\ref{fig:decon_timing}--\ref{fig:decon_proposal}. Figure~\ref{fig:decon_timing} shows the energy decrease over time in two difference scale. GA-fusion gives best performance among all tested algorithms. It is worthy of notice that ST-fusion gives poor performance. Some might expect better results with ST-fusion because they can achieve the optimal solution of the approximated function. However, tree approximation deletes too many edges ($\sim92\%$ of edges are deleted). To compare GA-proposal and ST-proposal, we generate 100 different approximated graphs of the Santa problem using our approach and another 100 using random spanning tree. We optimize former with $\alpha$-expansion and latter with dynamic programming. The results are plotted on Figure~\ref{fig:decon_approx}. Interestingly, the plot shows a curve rather than spread. Note that tree approximation requires $\sim92\%$ of edges to be deleted.

To figure out why our proposed method outperforms others, we provide more analysis while each graph cut-based algorithm is running (Fig.~\ref{fig:decon_proposal}). It reports the quality (energy) of the proposals and labeling ratio of each algorithm. According to section~\ref{sec:proposal}, ``good'' proposals satisfy the three conditions: quality, labeling rate, and diversity. First, GA-fusion produces the proposals with lower energy. It also achieves higher labeling rate than others. Finally, random jiggling of the plot implies that GA-fusion has very diverse proposals.

\renewcommand{\multirowsetup}{\centering}
\begin{table*}[t]
\caption{Image deconvolution results on four input images. Energies and average error rates are reported. The lowest energy for each case is in bold; GA-fusion achieves lowest energy for every image.}
\vspace{-3mm}
\begin{center}
\footnotesize
\begin{tabular}{lccccccc}
\hline
& GA-fusion & ST-fusion & $\alpha$-Expansion & Random-fusion & BP & TRW-S & MPLP-C \\
\hline
[Mean Energy ($\times 10^6$)]\\
Characters dataset & \textbf{-1.86} & 346.05 & 26.85 & 12.54 & 13.74 & 19.21 & 82.28 \\
White chessmen dataset & \textbf{-0.28} & 195.73 & 14.21 & 5.47 & 7.06 & 9.43 & 39.74 \\
Black chessmen dataset & \textbf{1.87} & 468.13 & 12.59 & 28.63 & 25.63 & 28.22 & 78.99 \\
\hline
[Average Error]\\
Characters dataset & \textbf{1.61$\%$} & 27.61$\%$ & 9.47$\%$ & 13.73$\%$ & 22.86$\%$ & 24.33$\%$ & 26.88$\%$ \\
White chessmen dataset & \textbf{0.63$\%$} & 19.63$\%$ & 7.59$\%$ & 9.07$\%$ & 16.77$\%$ & 17.94$\%$ & 20.29$\%$ \\
Black chessmen dataset & \textbf{2.33$\%$} & 65.90$\%$ & 7.71$\%$ & 37.42$\%$ & 61.84$\%$ & 63.74$\%$ & 65.79$\%$ \\
\hline
\end{tabular}
\end{center}
\label{table:deconvolution}
\end{table*}

\vspace{-1mm}
\subsubsection{Binary texture restoration}
\vspace{-1mm}

The aim of binary texture restoration is to reconstruct the original texture image from a noisy input. Although this problem has binary labels, move-making algorithms need to be applied because QPBO often fails and gives almost unlabeled solutions.

The energy function for texture restoration is formulated as the same as in~\cite{Cremers/ECCV06}. Unary cost is given by $\theta_p (x_p) = - \beta/(1+|I_p-x_p|)$, where $I_p$ is the color of the input image at pixel $p$, and $\beta$ is the unary cost weight. Pairwise costs are learned by computing joint histograms from the clean texture image. The costs for every edge within window size $w=35$ are learned first. Second, we choose a subset of edges to avoid overfitting. $S+N$ of most relevant edges are chosen, where $S$ is the number of submodular edges, and $N$ is the number of non-submodular edges. Relevance is given by the covariance of two nodes.

In the previous works, the numbers of edges $S$ and $N$ and the unary weight $\beta$ were determined by learning. However, the search space of the parameters was limited because they applied conventional graph cuts and QPBO. In~\cite{Cremers/ECCV06}, conventional graph cuts are used, thus $N$ should be fixed to zero. In~\cite{Kolmogorov/PAMI2007} QPBO is used to take account of non-submodular edges. However, QPBO gives almost unlabeled solutions when $N$ is large and $\beta$ is small.

To evaluate the capability of our algorithm, we control the model parameters so that each algorithm is applied on four different settings: low-connectivity and high-unary weight; low-connectivity and low-unary weight; high-connectivity and high-unary weight; and high-connectivity and low-unary weight. For low connectivity, we use six most relevant edges ($S=3$, $N=3$) and for high connectivity, we use 14 most relevant edges ($S=7$, $N=7$). The unary weight $\beta$ is chosen to be 5 and 20.

\begin{figure}[t]
\begin{center}
    \mbox{%
    \subfigure{ \includegraphics[width=0.2\linewidth]{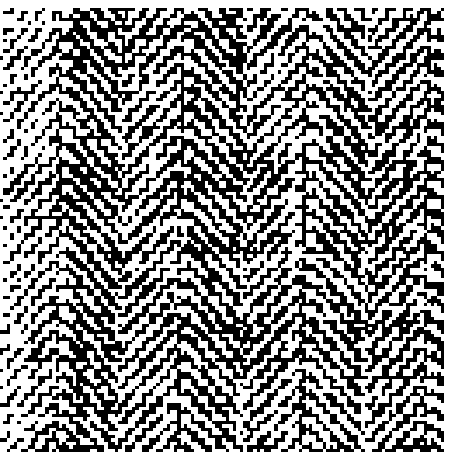}}
    \subfigure{ \includegraphics[width=0.2\linewidth]{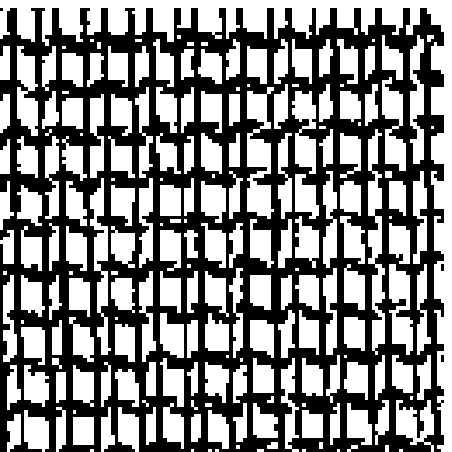}}
    \subfigure{ \includegraphics[width=0.2\linewidth]{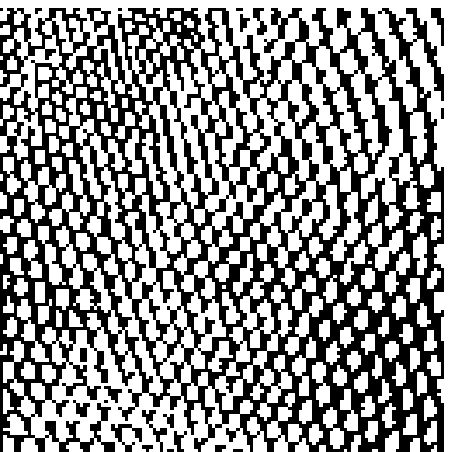}}
    \subfigure{ \includegraphics[width=0.2\linewidth]{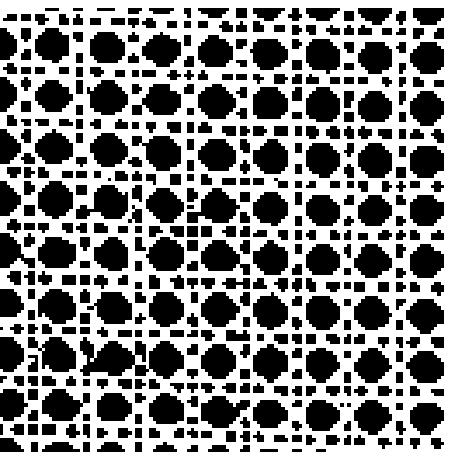}}
    }
\end{center}
\vspace{-3mm}
\caption{Four examples of Brodatz textures (cropped).
}
\label{fig:Brodatz}
\end{figure}

\begin{table*}[t]
\caption{Texture restoration experiments on 10 Brodatz textures. Average of relative energies is reported. Four different types of energy are considered by changing the number of pairwise costs and unary weight. The lowest energy for each case is in bold.}
\vspace{-3mm}
\begin{center}
\footnotesize
\begin{tabular}{lccccccc}
\hline
Energy type & QPBO & GA-fusion & ST-fusion & $\alpha$-expansion & random-fusion & BP &  TRW-S\\
\hline
\scriptsize{low-connectivity}  & \multirow{2}{*}{\textbf{0.0}} & \multirow{2}{*}{\textbf{0.0}} & \multirow{2}{*}{1.9} & \multirow{2}{*}{0.1} & \multirow{2}{*}{0.1} & \multirow{2}{*}{\textbf{0.0}} & \multirow{2}{*}{\textbf{0.0}} \\
\scriptsize{$\&$ high-unary weight}       &  &  &  &  &  &  &  \\
\hline
\scriptsize{low-connectivity}  & \multirow{2}{*}{n/a} & \multirow{2}{*}{\textbf{1.5}} & \multirow{2}{*}{25.6} & \multirow{2}{*}{2.8} & \multirow{2}{*}{5.1} & \multirow{2}{*}{3.6} & \multirow{2}{*}{10.6} \\
\scriptsize{$\&$ low-unary weight}      &  &  &   &  &  &  &  \\
\hline
\scriptsize{high-connectivity} & \multirow{2}{*}{n/a} & \multirow{2}{*}{0.9} & \multirow{2}{*}{25.3} & \multirow{2}{*}{\textbf{0.1}} & \multirow{2}{*}{0.9} & \multirow{2}{*}{0.4} & \multirow{2}{*}{3.3} \\
\scriptsize{$\&$ high-unary weight}       &  &  &   &  &  &  &  \\
\hline
\scriptsize{high-connectivity} & \multirow{2}{*}{n/a} & \multirow{2}{*}{\textbf{2.2}} & \multirow{2}{*}{38.3} & \multirow{2}{*}{8.3} & \multirow{2}{*}{4.6} & \multirow{2}{*}{6.0} & \multirow{2}{*}{11.6} \\
\scriptsize{$\&$ low-unary weight}      &  &   &  &  &  &  &  \\
\hline

\end{tabular}
\end{center}
\label{table:texture}
\end{table*}

For the input, we use the Brodatz texture dataset (Fig.~\ref{fig:Brodatz}), which contains different types of textures. Among them, 10 images are chosen for the purpose of this application. The chosen images have repeating patterns, and the size of the unit pattern is smaller than the window size (35-by-35). The images are resized to 256-by-256 pixels and binarized. Salt $\&$ pepper noise (70$\%$) is then added.

The results are summarized in Table~\ref{table:texture}. Relative energies\footnote{Relative
energy is calculated such that the energy of the best solution is 0 and that of
zero-labeled solution is 100.} are averaged over 10 texture images. When the problem is easy (low-connectivity and high-unary weight), QPBO is able to produce optimal solutions and all method except ST-fusion gives satisfactory low-energy results. Overall, GA-fusion consistently achieves low energy while others do not. QPBO and $\alpha$-expansion converged in 2.28 and 3.44 seconds on average, respectively. All other algorithms are iterated for 30 seconds.

\vspace{-1mm}
\subsection{Metric function optimization: OpenGM benchmark}
\vspace{-1mm}

We also evaluated our algorithm on metric functions. Some applications from OpenGM2 benchmark \cite{Kappes-2013-benchmark} are chosen: inpainting(n4), color segmentation(n4), and object segmentation. They all have Potts model for pairwise terms with 4-neighborhood structure. Since their energy functions are metric, the optimization is relatively easy compared to the previous energy functions.

The results are summarized in Table \ref{table:metric}. We report the average of final energies of GA-fusion, ST-fusion, and random-fusion after running 30 sec as well as other representative algorithms including $\alpha$-expansion, $\alpha\beta$-swap, BP, and TRW-S. Since these energy functions are relatively easy, the performance differences are not significant. Although GA-fusion aims to optimize non-metric functions, it is competitive to other algorithms and better than other heuristics such as ST-fusion and random-fusion.

\begin{table*}[t]
\caption{Mean energies obtained from deferent algorithms on metric energy functions. Test bed is from OpenGM2 benchmark. They all use the Potts model for designing pairwise terms.}
\vspace{-3mm}
\begin{center}
\footnotesize
\begin{tabular}{lrrrrrrr}
\hline
	&	GA-fusion	&	ST-fusion	&	$\alpha$-Expansion	&	$\alpha\beta$-Swap	&	Randon-fusion	&	BP	&	TRW-S	\\
\hline
Inpainting(n4)	&	\textbf{454.35}	&	466.92	&	\textbf{454.35}	&	454.75	&	545.96	&	\textbf{454.35}	&	490.48	\\
Color segmentation(n4)	&	20024.23	&	20139.12	&	20031.81	&	20049.9	&	24405.14	&	20094.03	&	\textbf{20012.18}	\\
Object segmentation	&	31323.07	&	31883.57	&	\textbf{31317.23}	&	31323.18	&	62834.62	&	35775.27	&	\textbf{31317.23}	\\
\hline
\end{tabular}
\end{center}
\label{table:metric}
\end{table*}

\vspace{-1mm}
\subsection{Synthetic function optimization}
\label{sec:synthetic}
\vspace{-1mm}

\begin{table}[t]
\caption{Energies obtained from deferent algorithms on synthetic problems. Test bed was designed to evaluate each algorithm on different ratios of non-metric term, coupling strengths $\lambda$, and connectivities. The name of the problem set indicates ``$\lambda$-(\textit{non-metric rate})-(\textit{graph structure})''. For each row, 10 results for different instances are averaged. The lowest energy for each case is in bold. GA-fusion consistently finds low energy solutions.}
\vspace{-6mm}
\begin{center}
\footnotesize
\begin{tabular}{lcccccc}
\hline
Energy type & GA & ST & $\alpha$-Exp & Rand & BP &  TRW-S\\
\hline
1-50-GRID4   & 0.1 & \textbf{0.0} & 90.7 & 0.3 & 5.7 & 6.8 \\
1-50-GRID8   & \textbf{0.4} & 2.0 & 100.0 & 1.5 & 9.0 & 17.1 \\
1-50-GRID24  & 1.1 & 33.2 & 100.0 & \textbf{0.3} & 11.5 & 17.2 \\
1-50-FULL    & \textbf{0.4} & 59.5 & 100.0 & 45.3 & 74.1 & 22.4 \\
\hline
10-50-GRID4  & 0.2 & \textbf{0.0} & 100.0 & 0.9 & 11.4 & 12.6 \\
10-50-GRID8  & \textbf{0.1} & 1.0 & 100.0 & 0.4 & 12.0 & 13.5 \\
10-50-GRID24 & 3.6 & 27.1 & 100.0 & \textbf{0.7} & 15.2 & 16.5 \\
10-50-FULL   & \textbf{0.0} & 1.7 & 100.0 & 5.2 & 1.8 & 1.8 \\
\hline
1-100-GRID8  & 0.3 & 0.8 & 100.0 & \textbf{0.0} & 1.2 & 2.0 \\
1-100-GRID24 & \textbf{0.0} & 0.9 & 100.0 & 3.0 & 2.6 & 2.5 \\
1-100-FULL   & \textbf{0.1} & 0.7 & 100.0 & 6.1 & 1.9 & 1.9 \\
\hline
10-100-GRID8 & 0.2 & 1.6 & 100.0 & \textbf{0.1} & 0.4 & 0.4 \\
10-100-GRID24& \textbf{0.0} & 2.2 & 100.0 & 3.1 & 2.2 & 1.8 \\
10-100-FULL  & \textbf{0.9} & 41.4 & 100.0 & 58.3 & 74.8 & 22.4 \\
\hline
\end{tabular}
\end{center}
\label{table:synthetic}
\end{table}

We compare our algorithm with others on various types of synthetic MRF problems to analyze performance further.

Four different types of graph structure are utilized: grid graphs with 4, 8, and 24 neighbors; and fully connected graph. The size of the grid graph is set to 30-by-30 and the size of the fully connected graph is 50. For each graph structure, we built five-label problems. Each unary cost is assigned by random sampling from uniform distribution: $\theta_p (x_p) \sim U(0,1)$. Pairwise costs are designed using the same method in section~\ref{sec:characteristics} (Equation~(\ref{eq:pair})). The difficulties of each problem are controlled by changing coupling strength $\lambda$ in the energy function~(\ref{eq:energy}). The amount of non-metric terms are set to 50$\%$ and 100$\%$ (non-metric term means pairwise cost which does not satisfy the condition~(\ref{eq:metric}))\footnote{For 4-neighborhood grid graph, 100$\%$ of non-metric terms are impossible because by simply flipping labels every term meets the condition~(\ref{eq:metric})}.
Ultimately, we construct 14 different types of MRF models, which are summarized in Table~\ref{table:synthetic} as ``$\lambda$-(\textit{non-metric rate})-(\textit{graph structure})''. For each type, 10 random instances of problems are generated.

Table~\ref{table:synthetic} reports the average of final energy from different algorithms. Some algorithms achieve low energy solutions with specific type of the energy function. GA-fusion consistently gives low energy solutions throughout all the energy type.

The following are some details on the experimental settings. Graph cut-based algorithms start from the zero-labeled initial. Every algorithm, except $\alpha$-expansion, is run for 10 sec because they do not follow a fixed rule for convergence. The experiment shows that 10 sec is enough time for every algorithm to converge. Although $\alpha$-expansion is fast, converging in less than a second, it mostly ended up with an zero-label. It is because that reduced sub-problem is too difficult and QPBO produces none of the labeled nodes in most cases.

\vspace{-1mm}
\section{Conclusions}
\label{sec:conclusions}
\vspace{-1mm}

Graph cuts-based algorithm is one of the most acclaimed algorithms for optimizing MRF energy functions. They can obtain the optimal solution for a submodular binary function and give a good approximation for multi-label function through the move-making approach. In the move-making approach, appropriate choice of the move space is crucial to performance. In other words, good proposal generation is required. However, efficient and generic proposals have not been available. Most works have relied on heuristic and application-specific ways. Thus, the present paper proposed a simple and application-independent way to generate proposals. With this proposal generation, we present a graph cuts-based move-making algorithm called GA-fusion, where the proposal is generated from approximated functions via graph approximation. We tested our algorithm on real and synthetic problems. Our experimental results show that our algorithm outperforms other methods, particularly when the problems are difficult.

\section*{Acknowledgments}
This research was supported in part by the Advanced Device Team, DMC R$\&$D, Samsung Electronics, and in part by the National Research Foundation of Korea (NRF) grant funded by the Ministry of Science, ICT $\&$ Future Planning (MSIP) (No. 2009-0083495).

{\small
\bibliographystyle{ieee}
\bibliography{mybib}
}

\end{document}